%% file: root.tex
\documentclass[letterpaper, 10 pt, conference]{IEEEconf}  

\IEEEoverridecommandlockouts                              
\usepackage{graphics} 
\usepackage{subcaption}
\usepackage{commath}
\usepackage{float}
\usepackage{soul}
\usepackage{epsfig} 
\usepackage{times} 
\usepackage{transparent}
\usepackage{amsmath} 
\usepackage{amssymb}  
\usepackage{amsfonts}   
\usepackage{cite}
\usepackage{bm}         
\usepackage{comment}
\usepackage[normalem]{ulem}
\usepackage{xcolor}
\usepackage{siunitx}
\newcommand\hem[1]{\textcolor{orange}{#1}}  
\graphicspath{{./}{images/}}
\usepackage{xspace}
\providecommand{\FullStop}{\text{~\@.\xspace}}
\providecommand{\Comma}{\text{~,\xspace}}       
\usepackage{todonotes}  %

\title{\LARGE \bf
Observer-based Controller Design for Oscillation Damping of a Novel Suspended Underactuated Aerial Platform
}

\author{Hemjyoti Das$^{1}$, 
Minh Nhat Vu$^{1,3}$, Tobias Egle
$^{1}$ and Christian Ott$^{1, 2}$
\thanks{$^{1}$Automation and Control Institute (ACIN), TU Wien, Gusshausstraße 27-29, 1040, Vienna, Austria
{\tt\small \{hemjyoti.das, minh.vu, tobias.egle,christian.ott\}@tuwien.ac.at} }%
\thanks{$^{2}$Institute of Robotics and Mechatronics, German Aerospace Center (DLR), Oberpfaffenhofen, Muenchener Strasse 20, 82234, Wessling, Germany}
\thanks{$^{3}$Austrian Institute of Technology (AIT) GmbH, 1210, Vienna, Austria}
}

\begin{document}

\maketitle
\thispagestyle{empty}
\pagestyle{empty}

\begin{abstract}
In this work, we present a novel actuation strategy for a suspended aerial platform. 
By utilizing an underactuation approach, we demonstrate the successful oscillation damping of the proposed platform, modeled as a spherical double pendulum. 
A state estimator is designed in order to obtain the deflection angles of the platform, which uses only onboard IMU measurements. 
The state estimator is an extended Kalman filter (EKF) with intermittent measurements obtained at different frequencies. 
An optimal state feedback controller and a PD+ controller are designed in order to dampen the oscillations of the platform in the joint space and task space respectively. 
The proposed underactuated platform is found to be more energy-efficient than an omnidirectional platform and requires fewer actuators. The effectiveness of our proposed system is validated using both simulations and experimental studies.
\end{abstract}

\section{Introduction}
\label{sec:intro}
Aerial robotic manipulation is a modern field of research that involves manipulation performed using a flying base \cite{ollero2021past}. In recent years, this field of aerial robotic manipulation has evolved significantly and has been utilized in numerous applications related to aerial inspection, construction, and load transportation  \cite{ollero2018aeroarms, ryll20176d, ollero2019aerial, ollero2021past, staub2018towards}. Research on aerial manipulation is often centered around an unmanned aerial vehicle (UAV) integrated with a robotic arm. In \cite{afifi2023physical, corsini2022nonlinear}, a fully actuated  UAV is mounted with a 3 degrees of freedom (DoF) robotic arm in order to perform collaborative human-handover tasks.  An impedance-based grasping task was demonstrated in \cite{huber2013first} using a UAV fitted with a 7-DoF Kuka robotic arm. 

The manipulator that can be mounted on the aerial vehicle for performing different tasks is often constrained by its weight.  For instance, an ultra lightweight arm \cite{suarez2023ultra} of 3.2 kg was developed and mounted on a medium-size multirotor platform in order to perform manipulation tasks on powerlines. However, in order to mount a 15 kg manipulator arm, a large scale helicopter system with a rotor-blade diameter of 3.7 meters was needed \cite{kondak2014aerial}. Such large dimensional rotors might be a concern for safety while operating in some complex environments. Moreover, the turbulence generated while operating in close proximity to the ground can be an additional cause of concern for the safety of the aerial system and its surroundings. A lot of energy is often consumed to compensate for gravity in aerial manipulator systems, which can limit the platform's payload capacity and its total time of flight (ToF). 

\begin{figure}
  \centering
  \def\svgwidth{0.7\columnwidth}
   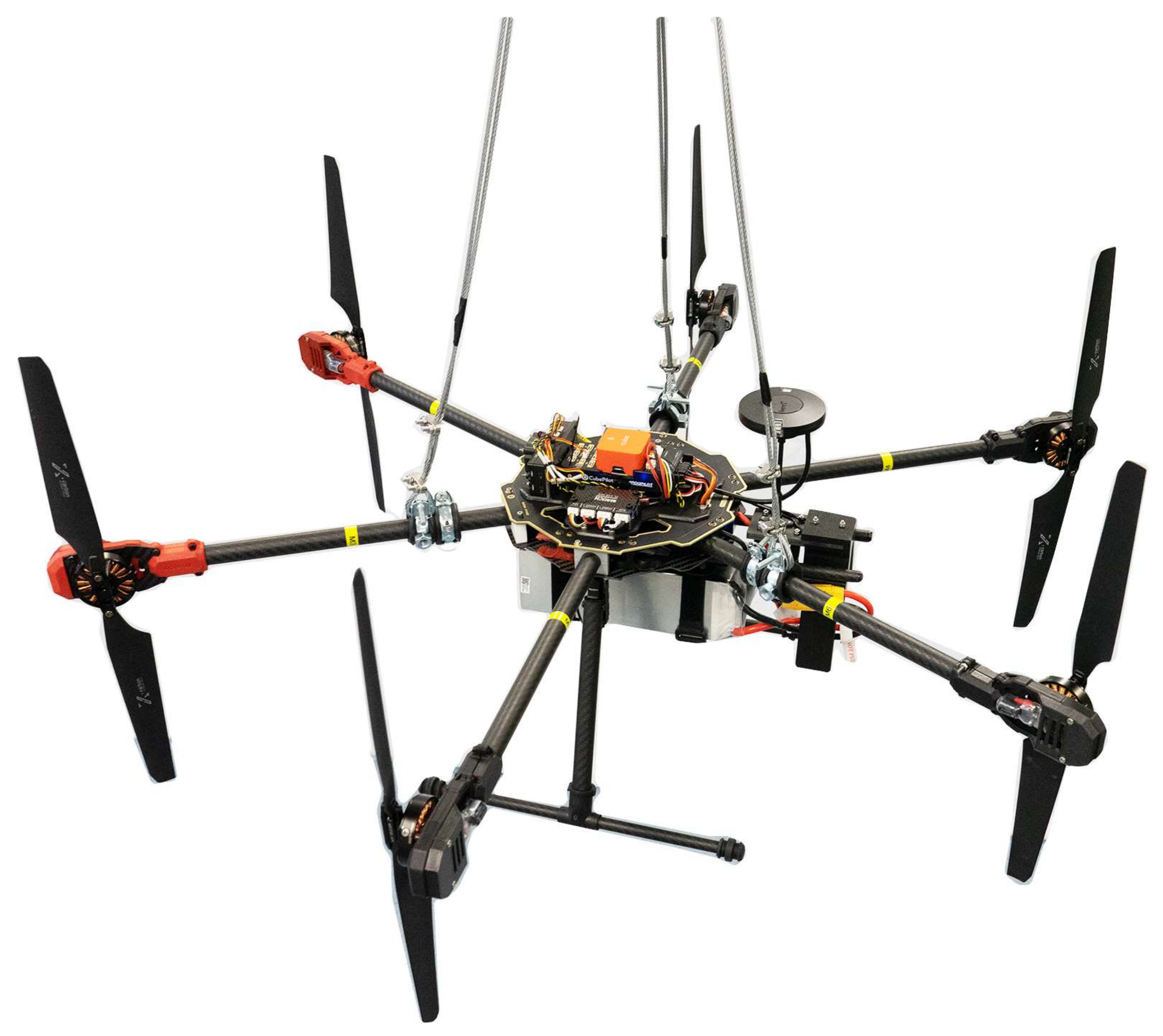
  \caption{Proposed planar-thrust 
  suspended aerial platform.}
  \label{fig:plat}
\end{figure}

    In order to mitigate the aforementioned issues, a suspended aerial manipulator (SAM) was proposed at the DLR \cite{sarkisov2019development}. It consists of an omnidirectional multirotor system equipped with a 7-DoF Kuka robotic arm, which is suspended from a carrier using a long cable. The gravity compensation due to the suspension reduces the energy consumption to a great extent and also ensures that the weight of the manipulator is not a limiting factor for its design. The SAM platform has been successfully utilized for demonstrating a number of hierarchical task-priority control applications \cite{coelho2021hierarchical, gabellieri2020compliance, sarkisov2023hierarchical}.  In \cite{yiugit2020preliminary}, a similar system was proposed which utilizes elastic suspension and can generate a 6-DoF wrench. This system was further augmented with a nonlinear model predictive control (NMPC) and a computed torque controller to assess its performance \cite{yiugit2021improving,yiugit2021novel}.\par
    
    One of the main challenges of such suspended aerial systems is the pendulum motion caused due to the suspension. The oscillation of the aerial base can arise due to its own motion, or due to the motion of the manipulator. Other factors such as wind gusts and other external disturbances can also contribute to its oscillatory motion. The control of oscillatory motion in free-flying multirotors with a cable-suspended load has been studied previously in \cite{sreenath2013geometric, zeng2019geometric, das2018dynamic, lv2022fixed}. However, the oscillatory motion in cable-suspended aerial platforms requires different damping techniques and has been studied in \cite{sarkisov2019development, sarkisov2020optimal}. In \cite{sarkisov2020optimal}, the double oscillations of the platform have been damped using optimal feedback control. It utilizes a simplified model of a planar double pendulum and considers the second oscillation angle to be approximately zero while framing the control law. Moreover, the platform used in \cite{sarkisov2019development} and \cite{sarkisov2023hierarchical} have eight propulsion units to achieve omni-directionality which consumes additional energy than an underactuated configuration. The damping control of suspended platforms in \cite{yiugit2020preliminary, yiugit2021improving, yiugit2021novel} utilizes exteroceptive sensor feedback using motion-capture systems, which is unfavorable for outdoor environments. \par
    
In this paper, we propose two different actuation concepts: a planar-thrust and a minimal-actuated configuration of a suspended aerial platform, which is the main contribution of this work.  These novel designs present a minimalist approach, as compared to some of the previous related works \cite{sarkisov2019development, sarkisov2023hierarchical}, \cite{yiugit2020preliminary}. By utilizing a minimalist approach, the proposed system is found to be more energy-efficient and it requires less number of actuators, while successfully damping the oscillations of the spherical double pendulum-type system.  A secondary contribution of this paper is the design of an extended Kalman filter (EKF) for the estimation of the oscillation angles of the suspended platform. The EKF relies on intermittent observations from its onboard sensors that are published at different frequencies. Finally, we discuss several control approaches for the proposed platforms. We design an optimal state feedback controller, which allows damping of the oscillations of the platform. Additionally, a PD+ controller is also designed in order to control the task-space dimensions. The proposed designs are evaluated using both computer simulations and experimental tests. \par 
The rest of the paper is organized as follows. Section \ref{sec:system} presents an overview of the suspended aerial system and explains its dynamics. The observer and control law are presented in Section \ref{sec:observer} and \ref{sec:controller}, respectively. The validation results are presented in Section \ref{sec:results}. Finally, the
conclusions and ideas for future work are given in Section \ref{sec:conclusion}.
\section{System Overview and Modelling}
\label{sec:system}
In this section, we first introduce the system dynamics, following which we review the omnidirectional suspended aerial platform in section \ref{sec:omni}. The novel underactuation approaches for the platform are then proposed in section \ref{sec:min_wrench} and \ref{sec:min_act}.
\subsection{System Dynamics}
\begin{figure}[t]
    \centering
    \def\svgwidth{0.5\columnwidth}
    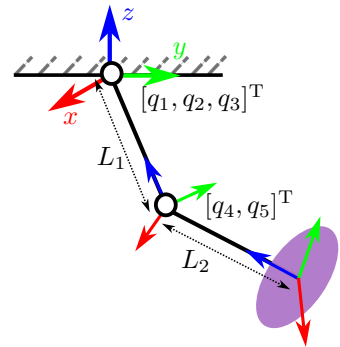
    \caption{Suspended aerial platform represented as a spherical double pendulum.   
    }%
    \label{fig:pendulum}
\end{figure}

We model the suspended platform as a spherical double pendulum as depicted in Fig. \ref{fig:pendulum}. The first spherical joint can be decomposed into three rotational joints $q_1$, $q_2$ and $q_3$, which are the rotation of the first link with length $L_1$ about its $x$, $y$, and $z$ axis, respectively. The second spherical joint can be expressed as the rotations $q_4$ and $q_5$ about its $x$ and $y$ axis, respectively, with the length of the second link denoted by $L_2$.  There is, however, no rotation about the $z$ axis for the second joint due to mechanical constraints imposed due to the suspension, while assuming all the cables to behave as rigid links. The double pendulum system has two modes of oscillation, which are the low-frequency high-amplitude mode for the first spherical joint, whereas the second joint is dominated by high-frequency low-amplitude mode \cite{sarkisov2019development, sarkisov2020optimal}.  
The joint dynamics can be summarized as
\begin{equation}
\label{eq:joint_dyn}
\mathbf{M(q)}\mathbf{\ddot{q}} + \mathbf{C(q, \dot{q})}\mathbf{\dot{q}} + \mathbf{g(q)} = \bm{\tau}\Comma
\end{equation} 
where $\mathbf{q} = \begin{bmatrix} q_1 & q_2 & q_3 & q_4 & q_5\end{bmatrix}^T$ is the configuration vector, $\mathbf{M(q)}$ is the inertia matrix and $\mathbf{C(q, \dot{q})\dot{q}}$ is the centrifugal/Coriolis term. $\mathbf{g(q)}$ is the gravity vector and $\bm{\tau}$ is the torque applied at the joints. The torques exerted at the joints $\bm{\tau}$ are related to the body wrenches $\mathbf{u}$ using the Jacobian $\mathbf{J(q)}$ as $\bm{\tau} = \mathbf{J(q)}^T \mathbf{u}$. Next, we transform these system dynamics to state-space form by defining the state vector as $\mathbf{x} = \begin{bmatrix} \mathbf{q} & \mathbf{\dot{q}} \end{bmatrix}^T$. 
The nonlinear state-transition matrix can be  then expressed as
   \begin{equation}
   \label{eq:non}
   \resizebox{0.91\hsize}{!}{$
   \mathbf{\dot{x}} =   \mathbf{f(x,u)} =   \begin{bmatrix}
    \mathbf{\dot{q}} \\
    \mathbf{M(q)^{-1}\left( \mathbf{J(q)}^T \mathbf{u}  - C(q,\dot{q})\dot{q} - g(q)\right)}
   \end{bmatrix}\FullStop
   $
   }
\end{equation} \par 
Note that our work concerns the analysis of only the suspended aerial base, and does not consider any mobile manipulation system which will be studied in the future.
\subsection{Omnidirectional Platform}
\label{sec:omni}
In an omnidirectional platform with unidirectional propellers \cite{tognon2018omnidirectional}, a 6 DoF wrench can be exerted in any direction with only a positive thrust vector from the propulsion unit. This wrench space includes the translational forces and the rotational moments about the center of the mass (COM) of the base. In the following, the forces along the $x$, $y$ and $z$ axis are represented as $F_x$, $F_y$ and $F_z$ respectively, while the rotational moments about the $x$, $y$ and $z$ axis are represented as  ${M_x}$,  ${M_y}$ and ${M_z}$ respectively.  The propulsion system used in this study is a combination of brushless DC motors and unidirectional propellers. The thrust vector generated by the motor can be expressed as $\mathbf{F}_{m}$. The wrench vector generated at the base, denoted as $\mathbf{u}_o$, can be related to the motor thrust using the allocation matrix $\mathbf{A}_o$ as $\mathbf{u}_o = \mathbf{A}_o \mathbf{F}_{m}$.
The allocation matrix $\mathbf{A}_o$ is obtained from the SAM platform\cite{sarkisov2019development} after scaling it down to the same size as our proposed platforms. The SAM has eight motors installed at an angle that is obtained by solving an optimization problem, which ensures a balanced design with equal distribution of wrench between the thrusters. The omnidirectional platform is used as a baseline for comparison with our proposed designs.
\subsection{Planar-thrust Platform}
\label{sec:min_wrench}
The planar-thrust platform (Fig. \ref{fig:plat}) has six propulsion units that can exert thrust only in the plane of its base. These thrusts can generate a wrench space that comprises the translational forces along its $x$ and $y$ axis, and moment about its $z$ axis. The  wrench vector $\mathbf{u_p}$
can be summarized as,
\begin{equation}
     \mathbf{u}_p =  \begin{bmatrix}
     {F_x} & {F_y}  & {M_z}
     \end{bmatrix}^{{T}}.
\end{equation} \par 
Even though this platform can exert only 3D wrench unlike a 6D wrench possible with platforms in \cite{sarkisov2019development, yiugit2020preliminary}, we still successfully demonstrate the oscillation-damping of the base in section \ref{sec:results}. Moreover, besides requiring fewer actuators than the omnidirectional design, we also show that such a planar-thrust design will allow us to conserve its actuation energy which serves as one of the main motivations for choosing such a design. The total translational force that can be exerted along the $x$ and $y$ axis increases for such a design as compared to an omnidirectional design. This might be beneficial for certain tasks that require the base to hold its position at desired deflected positions, while simultaneously damping its oscillations. The proposed planar-thrust design consists of six unidirectional thrusters, with the installation angle of each motor being $\pm$ 90 degrees. The direction of each thrust unit is highlighted by an arrow in Figure \ref{fig:plat_types}. The allocation-matrix $\mathbf{A}_p$ for this platform is given as
\begin{equation}
\label{eq:alloc_min_wrench}
    \mathbf{A}_p = \begin{bmatrix}
        1 & 1 & -0.5 & -0.5 & -0.5 & -0.5\\ 
        0 & 0 & 0.86 & 0.86 & -0.86 & -0.86 \\
        0.4 & -0.4 & 0.4 & -0.4 & -0.4 & 0.4 
    \end{bmatrix}\FullStop
\end{equation}
\subsection{Minimial-Actuated Platform}
\label{sec:min_act}
\begin{figure}[t]
    \centering
     \vspace{0.4cm}
    \def\svgwidth{0.7\columnwidth}
    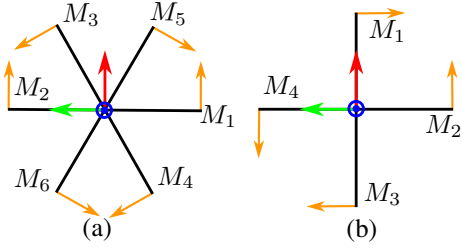
    \caption{Top-view of the proposed (a) planar-thrust platform with six rotors and  (b) minimal-actuated platform with four rotors. The arrows depict the direction of propulsion, whereas $M_i$ is used to denote the $i^{th}$ motor.}
    \label{fig:plat_types}%
\end{figure}

The concept of minimal-actuated design (Fig. \ref{fig:plat_types}) is similar to the planar-thrust design, as both of them have the same 3D wrench-space. However, the minimal-actuated platform consists of only four rotors, which is the minimum number required for covering the entire 3D wrench-space.  
It is to be noted that the actuation units utilize unidirectional propellers\footnote{Using a bidirectional propeller will allow us to further reduce the number of actuation units to three in order to achieve a 3D wrench, which will be studied in future.}. The allocation matrix  $\mathbf{A}_m$  for this platform is given as 

\begin{equation}
\label{eq:alloc_min_act}
   \mathbf{A}_m  = \begin{bmatrix}
        0 & 1 & 0 & -1 \\ 
        -1 & 0 & 1 & 0 \\
        -0.4 & 0.4 & -0.4 & 0.4 
    \end{bmatrix}\FullStop
\end{equation}
\section{Observer Design}
\label{sec:observer}
In this section, an extended Kalman filter (EKF) is designed in order to estimate the joint angles ${\mathbf{q}}$ and velocities $\dot{\mathbf{q}}$. The EKF is of the continuous-discrete form \cite{frogerais2011various} with intermittent measurements. The continuous-discrete version of the EKF is chosen because the measurements from the sensors are obtained at a discrete interval, while the process model in \eqref{eq:non} is a continuous-time differential equation. The prediction step for the states $\mathbf{x}$ and its error covariance $\mathbf{P}$ are as follows,
\begin{subequations}
\label{eq:ekf_pred}
\begin{align}
   \mathbf{\dot{{x}}}(t) &=   \mathbf{f}(\mathbf{x}(t),\mathbf{u}(t))\\
   \label{eq:cov}
       \mathbf{\dot{P}}(t) &= \mathbf{F}(t)\mathbf{P(t)} +\mathbf{P}(t) \mathbf{F}(t)^{T} + \mathbf{Q}(t)\Comma
\end{align}
\end{subequations}
where $\mathbf{F(t)}$ is the Jacobian of $\mathbf{f(t)}$ with respect to $\mathbf{x}$, while $\mathbf{Q(t)}$ is the process noise covariance. 
The measurement model used in the Kalman filter is given as,
\begin{equation}
    \label{eq:meas}
    \mathbf{z}_k = \mathbf{h(x}_k) + \mathbf{v}_k\Comma
\end{equation}
where the subscript $k$ denotes the sampling instance, $\mathbf{z}_k$ and $\mathbf{h(x}_k)$ are the measurement vector and measurement model respectively and $\mathbf{v}_k$ is the sensor noise covariance. The measurements used in the EKF are the translational velocity and the orientation of the suspended base, which are obtained directly from the flight-controller unit (FCU). The selection of these measurement quantities ensures the observability\cite{kreindler1964concepts} of our system states.
The optimal Kalman gain and the correction step for the state ${\mathbf{x}}$ and covariance $\mathbf{P}_k$ are presented as follows,
\begin{subequations}
    \label{eq:kalman_gain_update}
    \begin{align}
        \label{eq:kalman_gain_update a}
        \mathbf{K}_k &= \mathbf{P}_{k|k-1} \mathbf{H}_k^{T}\left(\mathbf{H}_k\mathbf{P}_{k|k-1}\mathbf{H}_k^T + \mathbf{R}_k \right)^{{-1}} \\
        \mathbf{{x}}(k|k) &=  \mathbf{{x}}(k|k-1) +   \mathbf{\lambda}_{i} \mathbf{K}_k\mathbf{(z}_k - \mathbf{h({x}}_{k|k-1}))\\
        \mathbf{P}_k &= (\mathbf{I} - \mathbf{K}_kH_k)\mathbf{{P}}(k|k-1)\Comma
    \end{align}
\end{subequations} 
where $\mathbf{R}_k$ is the covariance of the sensor noise $\mathbf{v}_k$ and  $\mathbf{H}_k$ is the derivative of the measurement model $\mathbf{h(x}_k)$ with respect to the states. For a particular sampling instant, if the ${i}_{th}$ sensor publishes a new measurement then the variable $\mathbf{\lambda}_{i}$ is one, and zero otherwise.
\section{Controller Design}
\label{sec:controller}
\subsection{Linear Quadratic Regulator (LQR)}
An optimal state feedback control LQR \cite{lewis1995optimal} is designed which minimizes the following linear quadratic cost function,
\begin{equation}
\label{eq:lqr}
   {J} =  \int_{0}^{t} \left ( \mathbf{x}\left ( t \right )^T\mathbf{\hat{Q}} \mathbf{x}\left ( t \right )  + \mathbf{u}\left(t\right )^T\mathbf{\hat{R}} \mathbf{u}\left ( t \right )\right )\mathrm{d}t\Comma \\
\end{equation}
where $\mathbf{\hat{Q}}$ is a positive semi-definite matrix which penalises the state $\mathbf{x}$ whereas the positive definite matrix $\mathbf{\hat{R}}$ penalizes the control input $\mathbf{u}$. 
The state feedback control law can be expressed as $\mathbf{u} = -\mathbf{kx}$, where the controller gain $\mathbf{k}$ is obtained by minimizing the cost function in \eqref{eq:lqr}. We choose the gain $\mathbf{k}$ by linearizing the system at its origin. The controllability matrix \cite{kreindler1964concepts} is found to be full rank, for both the omnidirectional and the proposed underactuated platforms, which conveys that the chosen wrench vector $\mathbf{u}_o$ and $\mathbf{u}_p$ are sufficient to control our system.
\subsection{PD+ Controller}
A PD+ control is designed in order to control the platform in the task space, which is described in this section.
\subsubsection{Omnidirectional Platform}
We consider the task coordinates  $\mathbf{x}_o$ of the omnidirectional platform as
\begin{equation}
    \label{eq:task_space_coor}
   \mathbf{x}_o = \begin{bmatrix}
        {p_x} & {p_y} & {\phi} & {\theta} & {\psi} 
    \end{bmatrix}^T,
\end{equation} 
where $p_x$ and $p_y$ are the inertial positions of the center of mass (COM) of the platform, while $\phi$, $\theta$ and $\psi$ are its Euler angles. The task-coordinates are related to the joints using the Jacobian $\mathbf{J}_o$ as $\dot{\mathbf{x}}_o = \mathbf{J}_o \dot{\mathbf{q}}$. Next, transforming the system dynamics \eqref{eq:joint_dyn}  to task space \cite{garofalo2018task}, we obtain
\begin{equation}
   \mathbf{ \Lambda} \ddot{\mathbf{x}}_o + \bm{\mu} \dot{\mathbf{x}}_o + \bm{\rho} = \mathbf{F}\Comma  
\end{equation}
where $\mathbf{\Lambda}$, $\bm{\mu}$, and $\bm{\rho}$ are the task-space inertia, Coriolis matrix, and gravity respectively. The control input $\mathbf{F}$ in task-space is then defined as\footnotemark{},  
\begin{equation}
\label{eq:task_space}
   \resizebox{0.95\hsize}{!}{
    $\mathbf{F} = \mathbf{\Lambda}\ddot{\mathbf{x}}_d + \bm{\mu} \dot{\mathbf{x}}_d + \mathbf{K}_d\left( \dot{\mathbf{x}}_d - \dot{\mathbf{x}}_o \right) + \mathbf{K}_p\left({\mathbf{x}}_d - \mathbf{x}_o \right) + \bm{\rho}$}\Comma
\end{equation}
where $\mathbf{K}_p$ and $\mathbf{K}_d$ are the respective coefficients corresponding to the position and velocity error in task space. $\mathbf{x}_d$, $\mathbf{\dot{x}}_d$ and $\mathbf{\ddot{x}}_d$ denote the desired task position, velocity and acceleration, respectively. By choosing control law \eqref{eq:task_space}, we obtain a stable closed-loop system dynamics as,
\begin{equation}
\label{eq:closed_loop}
\mathbf{\Lambda}(\ddot{\mathbf{x}}_d - \ddot{\mathbf{x}}_o) + \left( \bm{\mu} + \mathbf{K}_d  \right) \left( \dot{\mathbf{x}}_d - \dot{\mathbf{x}}_o\right) + \mathbf{K}_p\left({\mathbf{x}}_d - \mathbf{x}_o \right) = 0\FullStop
\end{equation} \par 
\footnotetext{If the task concerns damping the platform about the origin, then the compensation of gravity is not needed as it has a stabilizing effect.}
 The maximum thrust vector is denoted as $\mathbf{F}_{max}$, with the maximum thrust of each unit being limited to 9 N. In order to generate only positive thrust from the motors, a least-squares problem was solved to obtain the thrust ${\mathbf{F}_m}$ as,
\begin{equation}
\begin{aligned}
\min_{\mathbf{F}_m} \quad & \norm{\mathbf{F}_m}^2_2 \\
\textrm{s.t.} \quad  & \mathbf{J}^T\mathbf{A}_o\mathbf{F}_m=\mathbf{J}_o^T\mathbf{F}\Comma \\
& 0\leq \mathbf{F}_m\leq \mathbf{F}_{max}\FullStop    \\
\end{aligned}
\end{equation} 

\subsubsection{Planar-thrust and Minimial-actuated Plaforms}
Due to the reduced wrench-space, the task coordinates $\mathbf{x}_u$ of the underactuated platforms are chosen differently as follows, 
\begin{equation}
    \label{eq:task_space_coor}
   \mathbf{x}_u = \begin{bmatrix}
        {p_x} & {p_y} & {\psi} 
    \end{bmatrix}^T \FullStop
\end{equation} \par 
A redundancy is involved with this selection of task space as $\mathbf{x}_u \in \mathbb{R}^3$ whereas  $\mathbf{q} \in \mathbb{R}^5$. Therefore, we introduce the nullspace velocity $\mathbf{v}_n = \mathbf{N}\dot{\mathbf{q}}$ using the nullspace operator $\mathbf{N}$\cite{park1999dynamical}. The task coordinates are related to the joints using the inertial Jacobian $\mathbf{J}_u$ as $\dot{\mathbf{x}}_u = \mathbf{J}_u \dot{\mathbf{q}}$.
The system dynamics is then transformed into a decoupled task-space and nullspace coordinates \cite{garofalo2015inertially} as,
\begin{equation}
\label{eq:dyn_decop_vms}
\resizebox{1\hsize}{!}{
    $\begin{bmatrix}
       {\mathbf{\Lambda}}_x & 0 \\
        0 & {\mathbf{\Lambda}}_n
    \end{bmatrix}\begin{bmatrix}
        \ddot{{\mathbf{x}}}_u \\
        \dot{{\mathbf{v}}}_n
    \end{bmatrix} +
    \begin{bmatrix}
        {\bm{\mu}}_x & {\bm{\mu}}_{xn} \\
        {\bm{\mu}}_{nx} & {\bm{\mu}}_{n}
    \end{bmatrix}\begin{bmatrix}
        \dot{{\mathbf{x}}}_u \\
       {\mathbf{v}}_n
    \end{bmatrix} + \begin{bmatrix}
        {\bm{\rho}}_x \\
        {\bm{\rho}}_n
    \end{bmatrix}= \begin{bmatrix}
        {\mathbf{F}}_x\\
        {\mathbf{F}_n}
    \end{bmatrix}$,
}
\end{equation}
where the subscripts $n$ and $x$ refer to quantities in the nullspace and taskspace, respectively. In order to obtain stable closed-loop dynamics, the task-space controller $\mathbf{F}_x$ is chosen similar to \eqref{eq:task_space} as\footnotemark[\value{footnote}]
\begin{equation}
    \begin{aligned}
    \mathbf{F}_x =  &{\mathbf{\Lambda}}_e   \ddot{{\mathbf{x}}}_{d} + \bm{\mu}\dot{\mathbf{x}}_d + \mathbf{K}_d\left( \dot{\mathbf{x}}_{d} - \dot{\mathbf{x}}_u \right) + \mathbf{K}_p\left({\mathbf{x}_{d}} - \mathbf{x}_u \right)  \\
    & + \bm{\mu}_{xn} \bm{v}_n +   {\bm{\rho}}_x\FullStop
    \end{aligned}
\end{equation} \par 
The task-space force $\mathbf{F}_x$ and the nullspace force $\mathbf{F}_n$ are related to the joint torque $\bm{\tau}$ as follows,
\begin{equation}
    \bm{\tau}= \mathbf{J}_u^T\mathbf{F}_x +  \mathbf{N}^T\mathbf{F}_n \FullStop  
\end{equation} \par 
The torque $\bm{\tau}$ is related to the body wrenches $\mathbf{u}$ as  $\mathbf{J}^T\mathbf{u}$. Next, we define the relation $\mathbf{u} = \mathbf{B}\mathbf{u}_p$, where $\mathbf{u}_p$ consists of the selected wrench elements from the body-wrench $\mathbf{u}$, which the underactuated platform can exert and $\mathbf{B}$ is the underactuation matrix. The wrench component $\mathbf{u}_p$ is applied by actuation units of the platform, which is obtained using the following relation,
\begin{equation}
    \begin{bmatrix}
        \mathbf{u}_p \\ \mathbf{F}_n 
    \end{bmatrix} = \begin{bmatrix}
        \mathbf{J}^T\mathbf{B} & -\mathbf{N}^T
    \end{bmatrix}^{-1} \mathbf{J}_u^T\mathbf{F}_x \FullStop
\end{equation} \par 
Notice that the chosen actuation architecture not only generates the task force but also an additional force $\mathbf{F}_n$ in the nullspace dynamics.

\section{Results}
\label{sec:results}
Our proposed novel underactuated platforms are evaluated using both computer simulations and experimental studies, which will be discussed in this section. 
\subsection{Numerical Simulations}

\begin{figure*}[h]
    \centering
        \vspace{0.4cm}
    \includegraphics[width = 0.96\textwidth]{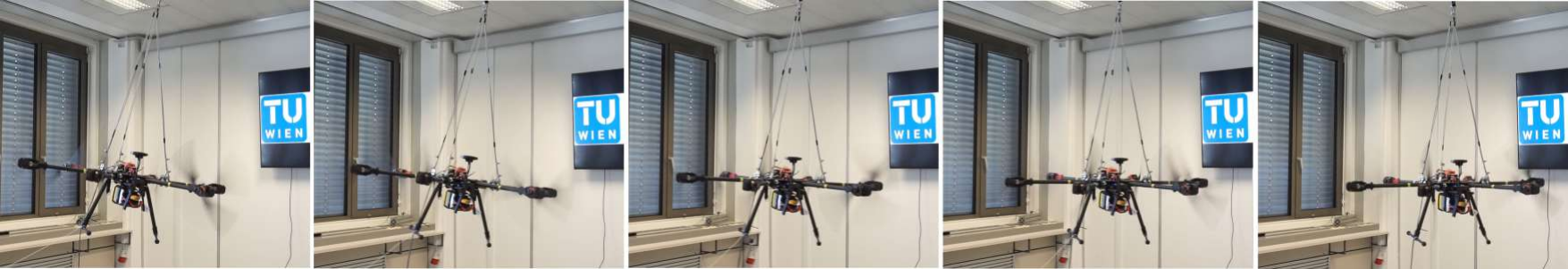}
    \caption{Sequence of the experiments with the proposed planar-thrust platform being stabilized using the LQR controller.}
    \label{fig:exp_sequence}
\end{figure*}

The simulation model used in our studies describes the spherical double pendulum in Fig. \ref{fig:pendulum}. The initial joint-angles for the simulation are chosen as $q_1$ = 0.15, $q_2$ = 0.2, $q_3$ = 0.2, $q_4$ = 0 and $q_5$ = 0 radians. The initial joint velocities are all considered as zero. The length of each link of the double pendulum is considered to be \SI{0.75}{\meter}, in order to replicate the actual experimental setup (Fig. \ref{fig:exp_sequence}). However, for outdoor experiments, the length of the first link will be higher than the second which will be studied in the future. The weight of the platform is found to be \SI{4.06}{\kilogram}, and its arm length is  \SI{0.4}{\meter}. The principal inertia along the $x$, $y$, and $z$ axis are found to be 0.0646, 0.0646, and 0.0682 $\SI{}{\kilogram\square\meter}$, respectively. 

In order to assume a similar comparison of the three different platforms, the state-penalty matrix $\mathbf{\hat{Q}}$ is chosen the same for all platforms as
\begin{equation*}
\resizebox{0.975\hsize}{!}{
    $\hat{\mathbf{Q}} = \mathrm{diag}{([200,200,20,0.01,0.01,50,50,1,0.01,0.01])}$}\FullStop
\end{equation*}\par
\begin{figure}[H]
    \centering
        \vspace{0.15cm}
    \includegraphics[width = 0.475\textwidth]{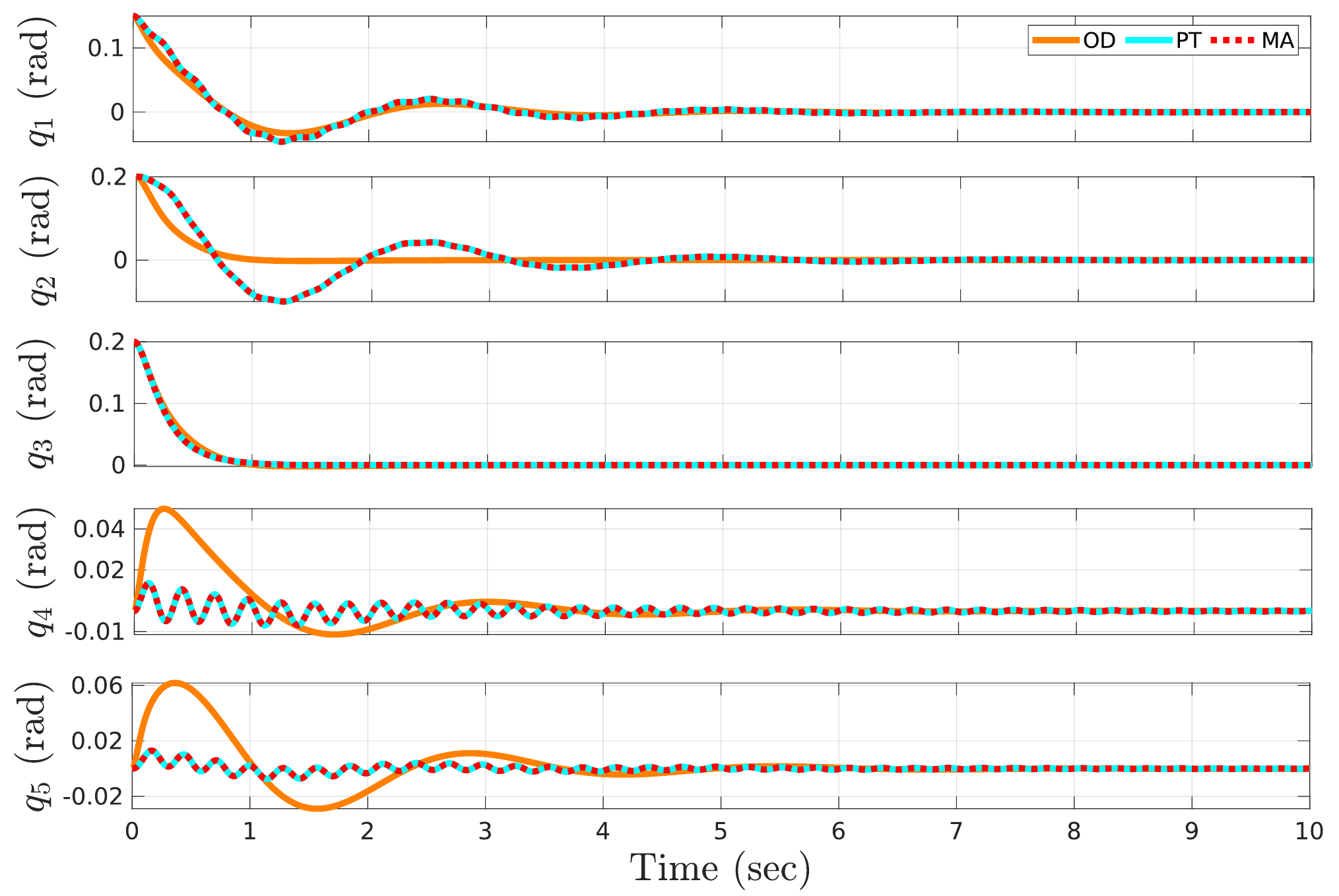}
    \caption{Joint-angles for the LQR-controlled system, using the omnidirectional platform (denoted as OD), planar-thrust platform (denoted as PT), and  minimum-actuated platform (denoted as MA).}
    \label{fig:sim_lqr_comp}
\end{figure}

The $\mathbf{\hat{R}}$ for all the platforms is chosen as an identity matrix. Moreover, the propulsion unit, mass, and inertia are considered to be the same for all three platforms. We observe that all three platforms have successfully damped the oscillations of the base, with a satisfactory settling time (Fig. \ref{fig:sim_lqr_comp}). However, for the second-spherical joint, the omnidirectional platform has a higher transient, as compared to the other platforms. The planar-thrust and the minimum-actuated platform demonstrate a similar behaviour for all five joints, with a high-frequency transient motion being observed for the joints $q_4$ and $q_5$,  before it converges to equilibrium.

\begin{figure}[H]
    \centering
    \includegraphics[width = 0.475\textwidth]{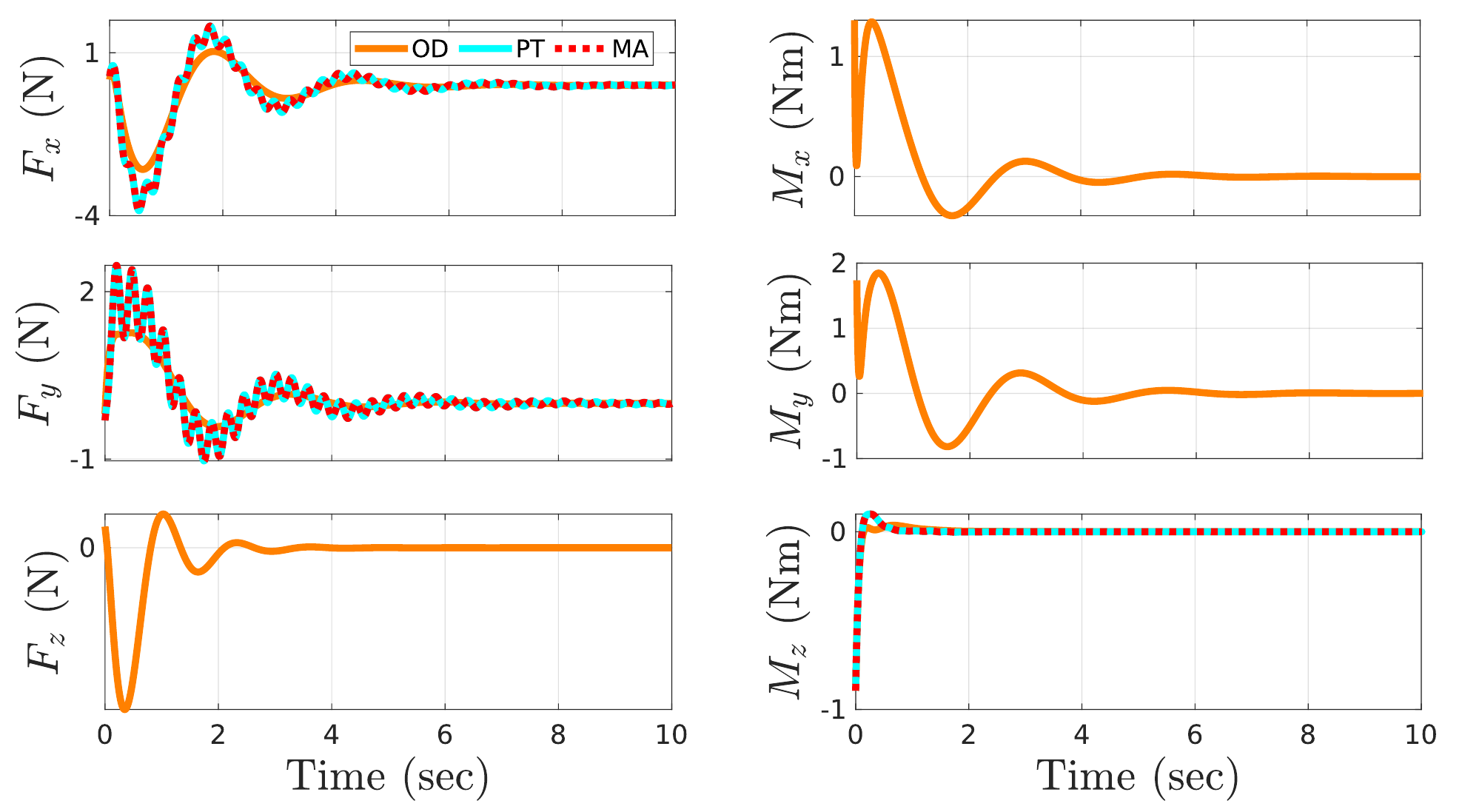}
    \caption{Wrenches commanded by the LQR controller.}
    \label{fig:sim_lqr_wrench_comp}
\end{figure}
\begin{figure}[H]
\centering
\begin{subfigure}[b]{0.475\textwidth}
   \includegraphics[width=1\linewidth]{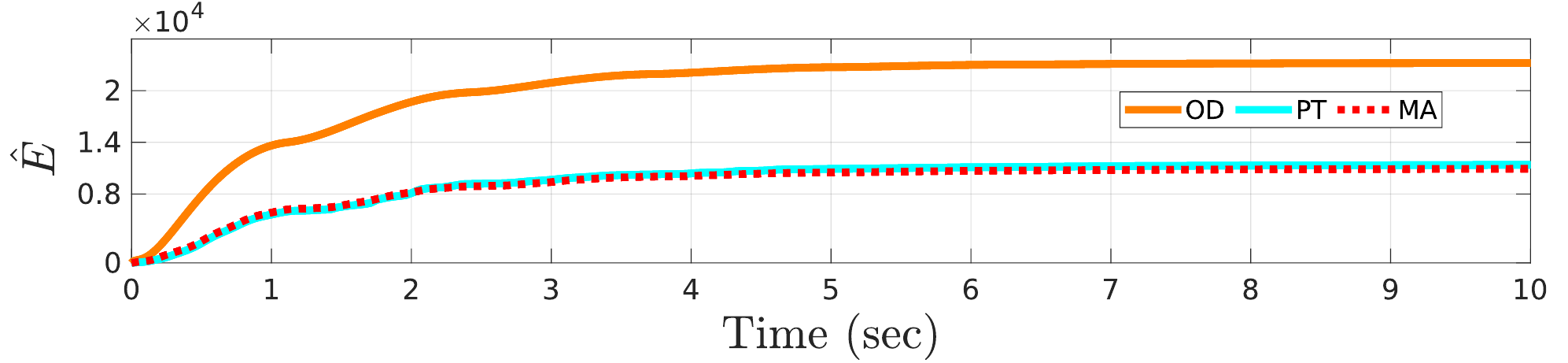}
   \caption{}
\end{subfigure}

\begin{subfigure}[b]{0.475\textwidth}
   \includegraphics[width=1\linewidth]{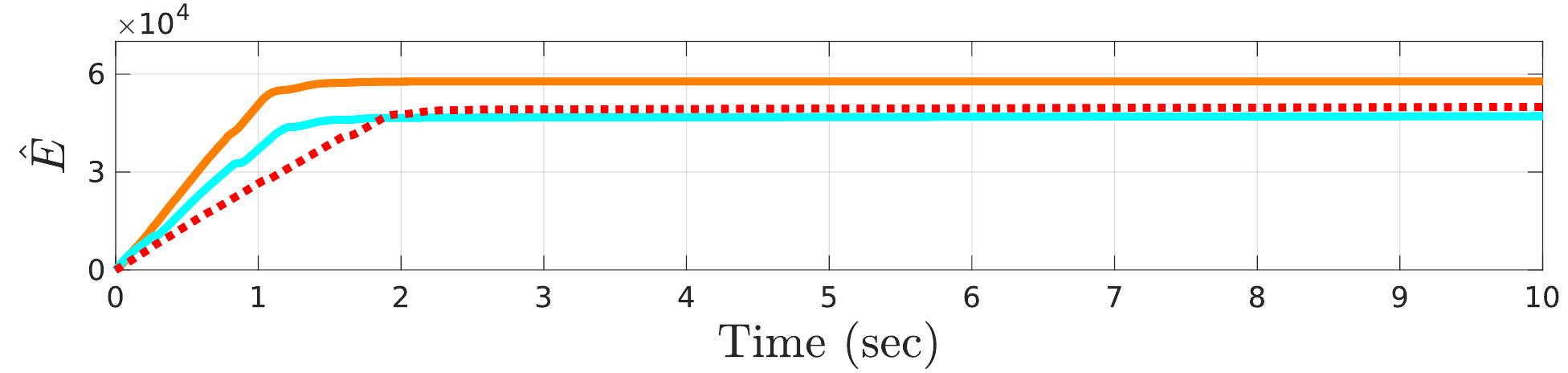}
   \caption{}
\end{subfigure}

\caption{Energy consumption for the (a) LQR-controlled system and (b) PD+ task-controlled system.}
\label{fig:energy}
\end{figure}

The wrenches commanded by the platform are shown in Fig. \ref{fig:sim_lqr_wrench_comp}. 
We observe that the translational forces $F_x$ and $F_y$ in the proposed planar-thrust and the minimum-actuated platforms have both high-frequency as well as low-frequency components, which is required to compensate for the two modes of the spherical double pendulum. The wrench exerted by the omnidirectional platform is predominantly of lower frequency which is sufficient to damp the motion of the base. In order to obtain an estimate of the energy consumption by the platforms, we introduce the quantity $\hat{E}$ as
\begin{equation}
    \hat{E} = \sum \left( \sum_{i=1}^{N}F_{m_i} . \mathrm{pwm}_i\right) \Delta t \Comma
\end{equation}
where $N$ is the number of actuators, $F_{m_i}$ is the thrust generated by $i^{th}$ motor and  $\text{pwm}_i$ is the PWM pulse issued to the motor $i$. The PWM pulse is used in computing $\hat{E}$ instead of the motor speed, because they are directly proportional to each other, besides the absence of motor speed sensing in the platform. $\hat{E}$ for the planar-thrust and the minimal-actuated platform is found to be lower than the omnidirectional platform by 50.9$\%$ and  52.8$\%$, respectively (Fig. \ref{fig:energy}).  \par 

\begin{figure}[h]
    \centering
    \includegraphics[width = 0.475\textwidth]{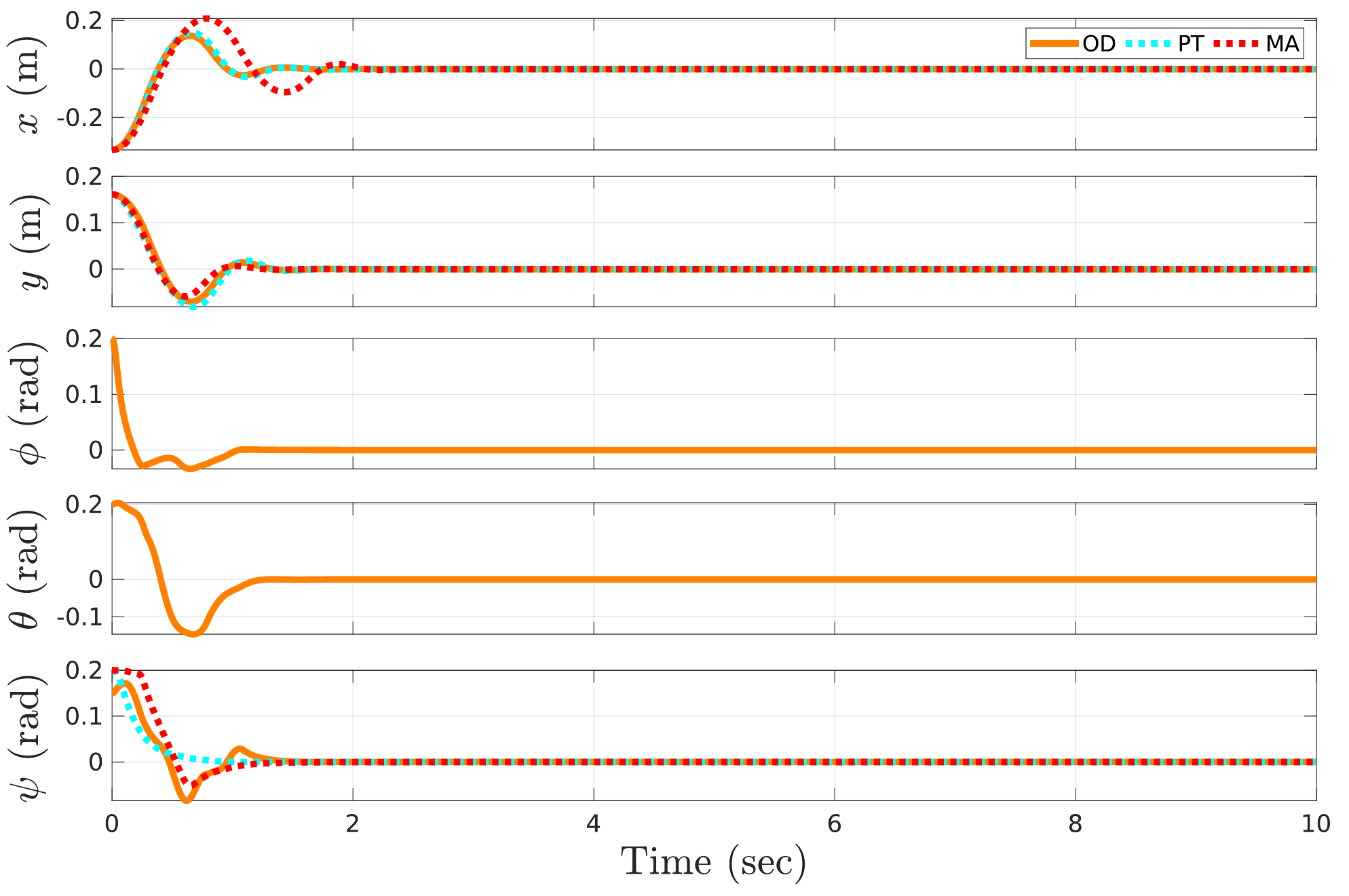}
    \caption{Task coordinates of the PD+ controlled system.}
    \label{fig:sim_task_space_comp}
\end{figure}
\begin{figure}[h]
    \centering
    \includegraphics[width = 0.475\textwidth]{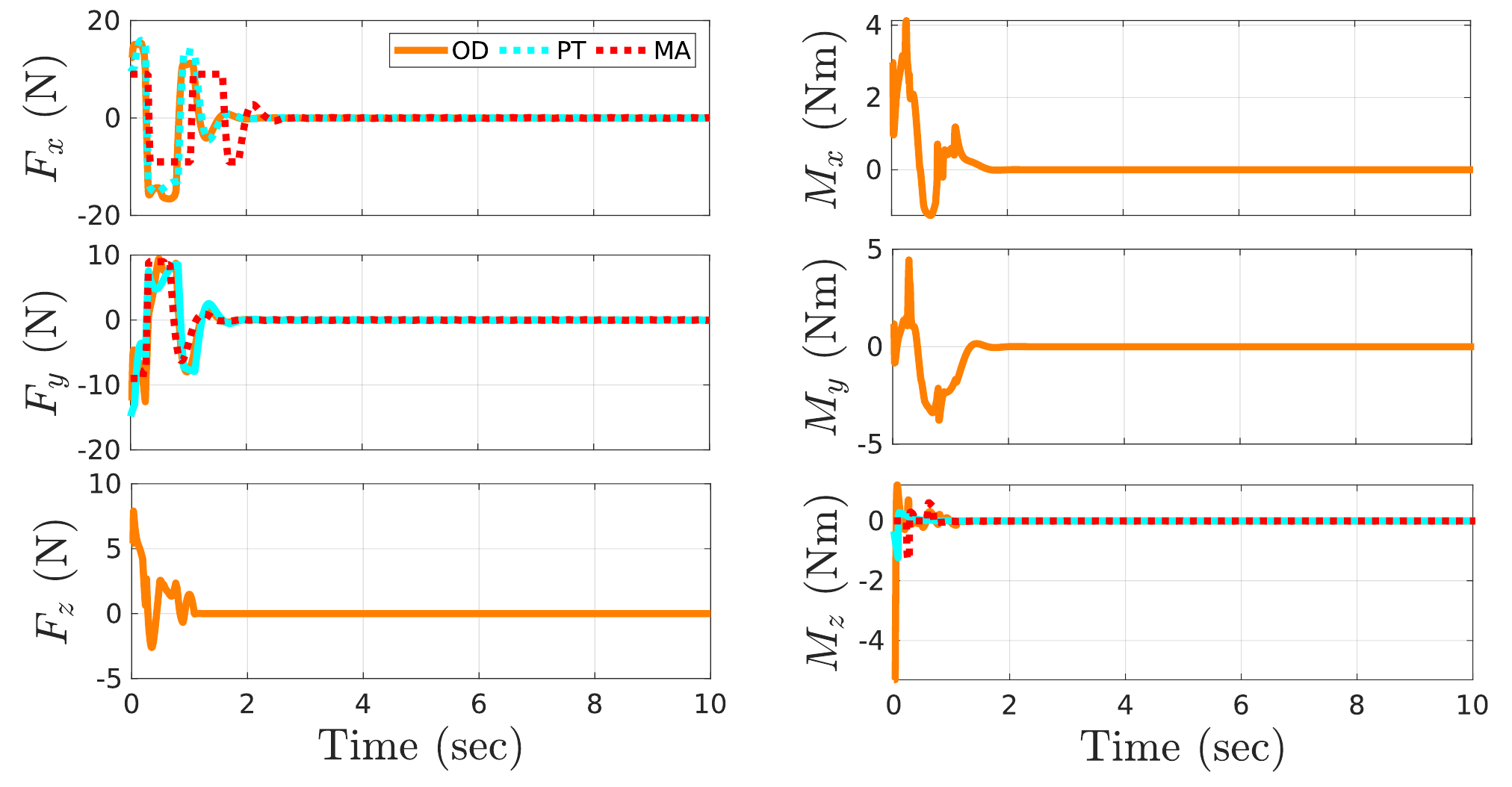}
    \caption{Wrenches commanded by the PD+ controller.}
    \label{fig:sim_task_wrench}
\end{figure}

The performance of the task-space controller and the corresponding commanded wrenches are shown in Fig. \ref{fig:sim_task_space_comp} and \ref{fig:sim_task_wrench}, respectively. In order to assume a similar comparison for the different platforms, the same stiffness $\mathbf{K}_p$ is chosen for all of them. The $\mathbf{K}_p$ corresponding to the states $x$, $y$, and $\psi$ are chosen as 400, 400, and 100, respectively. The fully-actuated platform has two additional task spaces $\phi$ and $\theta$, and their corresponding $\mathbf{K}_p$ were chosen as 100. The damping coefficient $\mathbf{K}_d$ is chosen as twice the square root of $\mathbf{K}_p$. The performance for all the platforms is similar in terms of the state's evaluation. It is also observed that similar to the previous case, the proposed platforms were more energy-efficient than the omnidirectional configuration (Fig. \ref{fig:energy}).
\subsection{Experimental Results}
\begin{figure}[h]
    \centering
    \includegraphics[width = 0.475\textwidth]{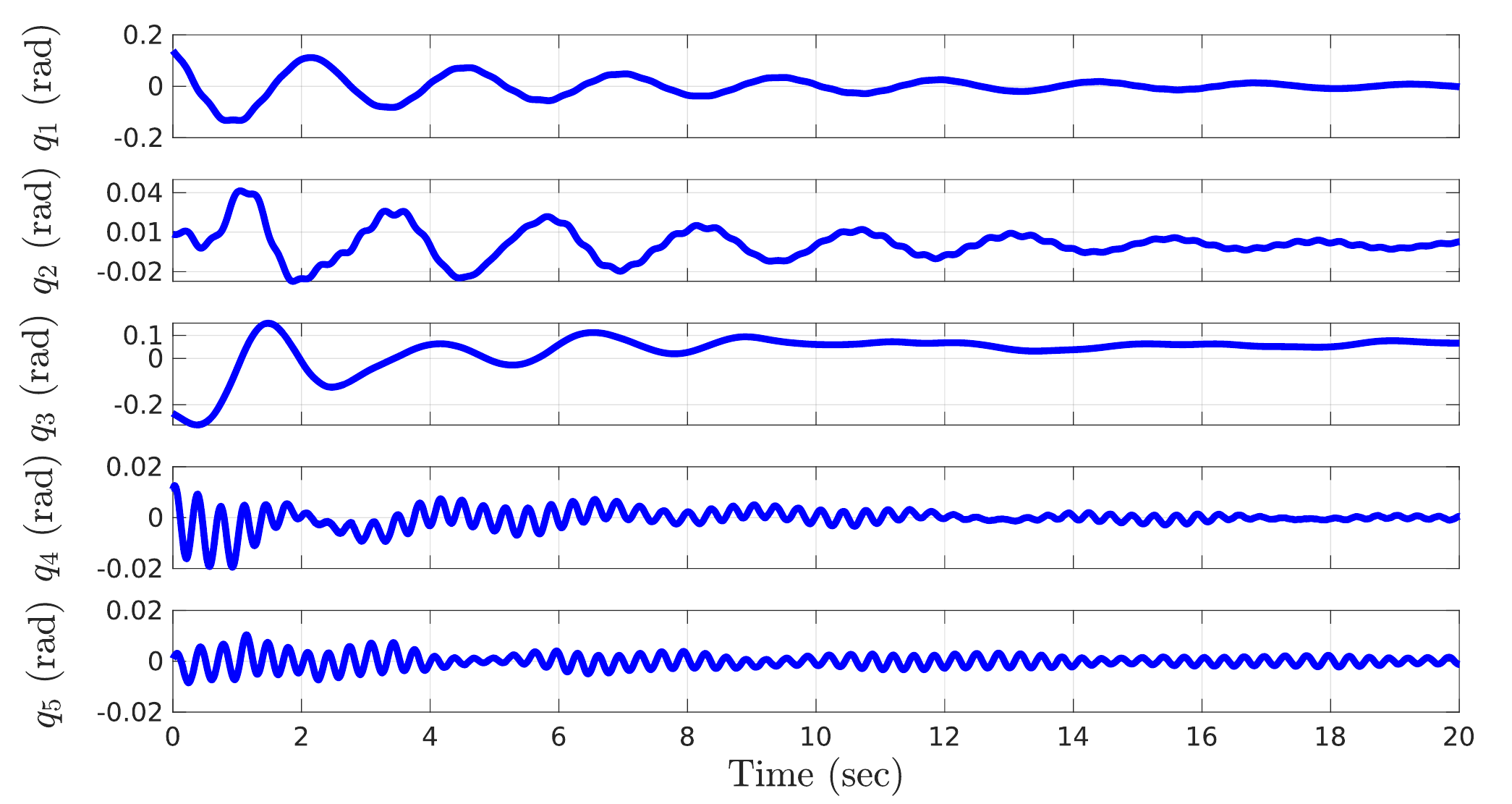}
    \caption{Experimental results showing the joint-angles for the LQR-controlled system using the planar-thrust platform.}
    \label{fig:expm_lqr_comp}
\end{figure}
\begin{figure}[h]
    \centering
    \includegraphics[width = 0.475\textwidth]{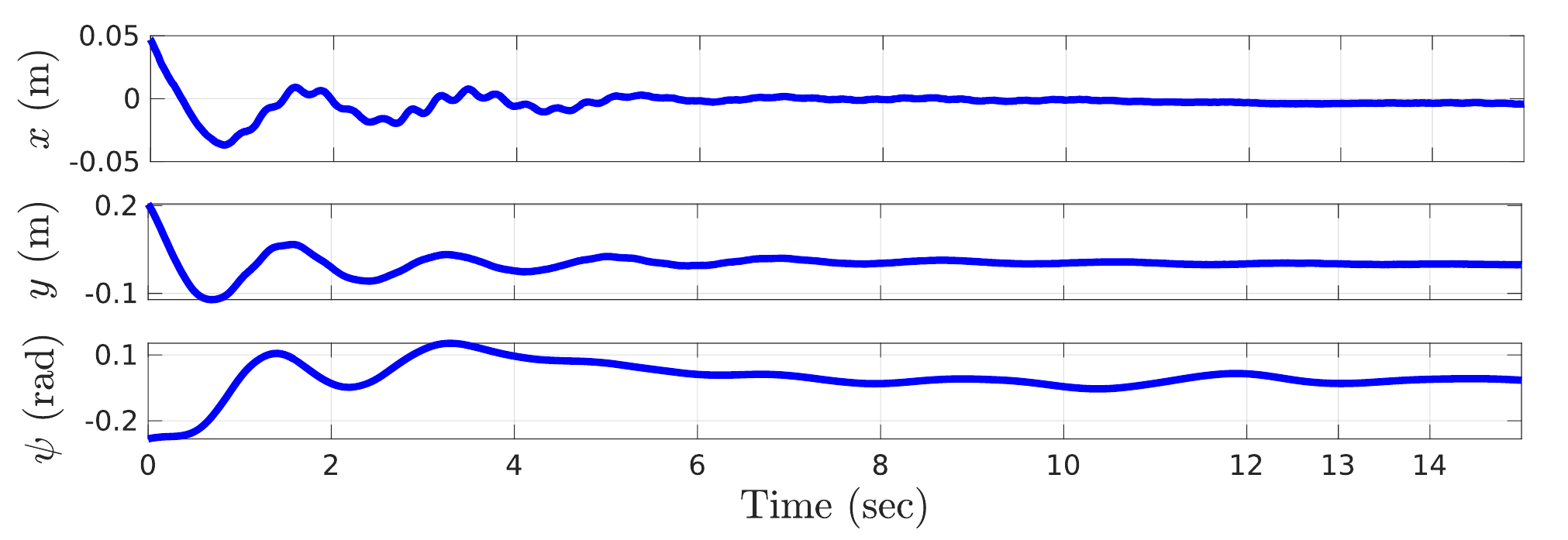}
    \caption{Experimental results for the planar-thrust platform with the PD+ task-space controller.}
    \label{fig:expm_task_space_comp}
\end{figure}
The experiments are conducted using the planar-thrust platform (Fig. \ref{fig:exp_sequence}). It comprises six brushless DC motors with a rating of 380 KV and folding propellers with a diameter of 15 inches. Such a configuration of the propulsion system is sufficient for our medium-size platform. The flight controller unit (FCU) is chosen as the Pixhawk Cube Orange which is lightweight and compact and includes three IMU sensors for additional accuracy of our estimation algorithm. We utilize the PX4 firmware \cite{meier2015px4} in order to interface the FCU, and our algorithm runs at a frequency of 500 Hz. \par 

The experiments demonstrate the damping behaviour of the proposed planar-thrust platform by stabilizing it from an initial tilt angle (Fig. \ref{fig:exp_sequence}). 
For the LQR-controller, $\hat{\mathbf{R}}$ is chosen as identity matrix, while  $\hat{\mathbf{Q}}$ is chosen as 
\begin{equation*}
\resizebox{0.975\hsize}{!}{
    $\hat{\mathbf{Q}} = \mathrm{diag}{([100,100,0.1,0.01,0.01,1,1,0.1,0.0001,0.0001])}$}\FullStop
\end{equation*}\par
A high-frequency oscillation for the joints $q_4$ and $q_5$ with an amplitude of around 0.01 radian is observed (Fig. \ref{fig:expm_lqr_comp}). A similar phenomenon was also observed in the simulations (Fig. \ref{fig:sim_lqr_comp}). This is possibly due to the low bandwidth of the motor to counteract high-frequency oscillations. For the task-space controlled system, $\mathbf{K}_p$ corresponding to the states $x$, $y$, and $\psi$ are chosen as 25, 25, and 0.35 respectively, while their corresponding $\mathbf{K}_d$ are chosen as 10, 10 and 0.13. We observe that the states are damped within 8 seconds (Fig. \ref{fig:expm_task_space_comp}). However, there is a time delay in sensing the yaw angle $\psi$, which resulted in its slower convergence rate, as compared to the other states.

\section{Conclusion and Future Work}
\label{sec:conclusion}
In this paper, we proposed a novel planar-thrust and a minimum-actuated suspended aerial platform and compared their performance with an omnidirectional design. The proposed designs were found to be more energy-efficient and use fewer actuators than the omnidirectional configuration, while successfully damping the spherical double pendulum-type system. In order to validate and compare the platforms, we designed optimal state-feedback and a PD+ controller for damping the oscillations of the platform in the joint space and task space, respectively. An intermittent extended Kalman filter was designed using only its onboard IMU measurements, in order to provide state feedback to the controllers. \par 
In the future, we plan to mount a robotic manipulator on the platform base and perform aerial manipulation tasks with it. We would like to include additional sensors in order to obtain fast and drift-free reliable estimates of the yaw angle. It would also be interesting to analyse the use of bi-directional propellers to further reduce the number of actuators for the minimal configuration. We would also like to introduce learning-based algorithms in order to improve the system modeling while accounting for unknown disturbances, which will further improve the tracking by the controller.

\bibliographystyle{IEEEtran}
\bibliography{bibliography}{}
\end{document}

%% file: images/drone-drawing-new.pdf_tex
\begingroup%
  \makeatletter%
  \providecommand\color[2][]{%
    \errmessage{(Inkscape) Color is used for the text in Inkscape, but the package 'color.sty' is not loaded}%
    \renewcommand\color[2][]{}%
  }%
  \providecommand\transparent[1]{%
    \errmessage{(Inkscape) Transparency is used (non-zero) for the text in Inkscape, but the package 'transparent.sty' is not loaded}%
    \renewcommand\transparent[1]{}%
  }%
  \providecommand\rotatebox[2]{#2}%
  \newcommand*\fsize{\dimexpr\f@size pt\relax}%
  \newcommand*\lineheight[1]{\fontsize{\fsize}{#1\fsize}\selectfont}%
  \ifx\svgwidth\undefined%
    \setlength{\unitlength}{1212.94460507bp}%
    \ifx\svgscale\undefined%
      \relax%
    \else%
      \setlength{\unitlength}{\unitlength * \real{\svgscale}}%
    \fi%
  \else%
    \setlength{\unitlength}{\svgwidth}%
  \fi%
  \global\let\svgwidth\undefined%
  \global\let\svgscale\undefined%
  \makeatother%
  \begin{picture}(1,0.89118117)%
    \lineheight{1}%
    \setlength\tabcolsep{0pt}%
    \put(0,0){\includegraphics[width=\unitlength,page=1]{drone-drawing-new.pdf}}%
    \put(0.51780358,0.20546476){\color[rgb]{0,0,0}\makebox(0,0)[lt]{\lineheight{1.25}\smash{\begin{tabular}[t]{l}\scalebox{0.9}{Propulsion Unit}\end{tabular}}}}%
    \put(0.7004514,0.77551734){\color[rgb]{0,0,0}\makebox(0,0)[lt]{\lineheight{1.25}\smash{\begin{tabular}[t]{l}\scalebox{0.9}{Flight Controller Unit}\end{tabular}}}}%
    \put(0.00423745,0.77087191){\color[rgb]{0,0,0}\makebox(0,0)[lt]{\lineheight{1.25}\smash{\begin{tabular}[t]{l}\scalebox{0.9}{Suspension Cables}\end{tabular}}}}%
    \put(0,0){\includegraphics[width=\unitlength,page=2]{drone-drawing-new.pdf}}%
  \end{picture}%
\endgroup%

%% file: images/schematics.pdf_tex
\begingroup%
  \makeatletter%
  \providecommand\color[2][]{%
    \errmessage{(Inkscape) Color is used for the text in Inkscape, but the package 'color.sty' is not loaded}%
    \renewcommand\color[2][]{}%
  }%
  \providecommand\transparent[1]{%
    \errmessage{(Inkscape) Transparency is used (non-zero) for the text in Inkscape, but the package 'transparent.sty' is not loaded}%
    \renewcommand\transparent[1]{}%
  }%
  \providecommand\rotatebox[2]{#2}%
  \newcommand*\fsize{\dimexpr\f@size pt\relax}%
  \newcommand*\lineheight[1]{\fontsize{\fsize}{#1\fsize}\selectfont}%
  \ifx\svgwidth\undefined%
    \setlength{\unitlength}{157.30660386bp}%
    \ifx\svgscale\undefined%
      \relax%
    \else%
      \setlength{\unitlength}{\unitlength * \real{\svgscale}}%
    \fi%
  \else%
    \setlength{\unitlength}{\svgwidth}%
  \fi%
  \global\let\svgwidth\undefined%
  \global\let\svgscale\undefined%
  \makeatother%
  \begin{picture}(1,1.15544667)%
    \lineheight{1}%
    \setlength\tabcolsep{0pt}%
    \put(0,0){\includegraphics[width=\unitlength,page=1]{schematics.pdf}}%
    \put(0.38857547,0.7476539){\color[rgb]{0,0,0}\makebox(0,0)[lt]{\lineheight{1.25}\smash{\begin{tabular}[t]{l}$[q_1,q_2,q_3]^\mathrm{T}$\end{tabular}}}}%
    \put(0.25697109,0.56541899){\color[rgb]{0,0,0}\makebox(0,0)[lt]{\lineheight{1.25}\smash{\begin{tabular}[t]{l}$L_1$\end{tabular}}}}%
    \put(0.51568843,0.2479039){\color[rgb]{0,0,0}\makebox(0,0)[lt]{\lineheight{1.25}\smash{\begin{tabular}[t]{l}$L_2$\end{tabular}}}}%
    \put(0.33252219,1.01282343){\color[rgb]{0,0,0}\makebox(0,0)[lt]{\lineheight{1.25}\smash{\begin{tabular}[t]{l}\textcolor{blue}{$z$}\end{tabular}}}}%
    \put(0.1525503,0.6934171){\color[rgb]{0,0,0}\makebox(0,0)[lt]{\lineheight{1.25}\smash{\begin{tabular}[t]{l}\textcolor{red}{$x$}\end{tabular}}}}%
    \put(0.49059519,0.90807069){\color[rgb]{0,0,0}\makebox(0,0)[lt]{\lineheight{1.25}\smash{\begin{tabular}[t]{l}\textcolor{green}{$y$}\end{tabular}}}}%
    \put(0.58811658,0.41926779){\color[rgb]{0,0,0}\makebox(0,0)[lt]{\lineheight{1.25}\smash{\begin{tabular}[t]{l}$[q_4,q_5]^\mathrm{T}$\end{tabular}}}}%
    \put(0,0){\includegraphics[width=\unitlength,page=2]{schematics.pdf}}%
  \end{picture}%
\endgroup%

%% file: images/underactuated_design.pdf_tex
\begingroup%
  \makeatletter%
  \providecommand\color[2][]{%
    \errmessage{(Inkscape) Color is used for the text in Inkscape, but the package 'color.sty' is not loaded}%
    \renewcommand\color[2][]{}%
  }%
  \providecommand\transparent[1]{%
    \errmessage{(Inkscape) Transparency is used (non-zero) for the text in Inkscape, but the package 'transparent.sty' is not loaded}%
    \renewcommand\transparent[1]{}%
  }%
  \providecommand\rotatebox[2]{#2}%
  \newcommand*\fsize{\dimexpr\f@size pt\relax}%
  \newcommand*\lineheight[1]{\fontsize{\fsize}{#1\fsize}\selectfont}%
  \ifx\svgwidth\undefined%
    \setlength{\unitlength}{644.31016817bp}%
    \ifx\svgscale\undefined%
      \relax%
    \else%
      \setlength{\unitlength}{\unitlength * \real{\svgscale}}%
    \fi%
  \else%
    \setlength{\unitlength}{\svgwidth}%
  \fi%
  \global\let\svgwidth\undefined%
  \global\let\svgscale\undefined%
  \makeatother%
  \begin{picture}(1,0.50887067)%
    \lineheight{1}%
    \setlength\tabcolsep{0pt}%
    \put(0,0){\includegraphics[width=\unitlength,page=1]{underactuated_design.pdf}}%
    \put(0.17558879,0.00644965){\color[rgb]{0,0,0}\makebox(0,0)[lt]{\lineheight{1.25}\smash{\begin{tabular}[t]{l}(a)\end{tabular}}}}%
    \put(0.80449077,0.44301324){\color[rgb]{0,0,0}\makebox(0,0)[lt]{\lineheight{1.25}\smash{\begin{tabular}[t]{l}$M_1$\end{tabular}}}}%
    \put(0.92893376,0.23411395){\color[rgb]{0,0,0}\makebox(0,0)[lt]{\lineheight{1.25}\smash{\begin{tabular}[t]{l}$M_2$\end{tabular}}}}%
    \put(0.7942373,0.0847354){\color[rgb]{0,0,0}\makebox(0,0)[lt]{\lineheight{1.25}\smash{\begin{tabular}[t]{l}$M_3$\end{tabular}}}}%
    \put(0.57232513,0.31855654){\color[rgb]{0,0,0}\makebox(0,0)[lt]{\lineheight{1.25}\smash{\begin{tabular}[t]{l}$M_4$\end{tabular}}}}%
    \put(0.02440276,0.11526008){\color[rgb]{0,0,0}\rotatebox{-0.68839021}{\makebox(0,0)[lt]{\lineheight{1.25}\smash{\begin{tabular}[t]{l}$M_6$\end{tabular}}}}}%
    \put(0,0){\includegraphics[width=\unitlength,page=2]{underactuated_design.pdf}}%
    \put(0.13159152,0.47693393){\color[rgb]{0,0,0}\rotatebox{1.7810289}{\makebox(0,0)[lt]{\lineheight{1.25}\smash{\begin{tabular}[t]{l}$M_3$\end{tabular}}}}}%
    \put(0.32289974,0.48144327){\color[rgb]{0,0,0}\rotatebox{-1.1282704}{\makebox(0,0)[lt]{\lineheight{1.25}\smash{\begin{tabular}[t]{l}$M_5$\end{tabular}}}}}%
    \put(0.43258009,0.25519665){\color[rgb]{0,0,0}\rotatebox{-0.86609794}{\makebox(0,0)[lt]{\lineheight{1.25}\smash{\begin{tabular}[t]{l}$M_1$\end{tabular}}}}}%
    \put(0.33935259,0.1199702){\color[rgb]{0,0,0}\rotatebox{0.25543067}{\makebox(0,0)[lt]{\lineheight{1.25}\smash{\begin{tabular}[t]{l}$M_4$\end{tabular}}}}}%
    \put(0.02336708,0.31731141){\color[rgb]{0,0,0}\rotatebox{-1.5082177}{\makebox(0,0)[lt]{\lineheight{1.25}\smash{\begin{tabular}[t]{l}$M_2$\end{tabular}}}}}%
    \put(0.75624733,0.00525639){\color[rgb]{0,0,0}\makebox(0,0)[lt]{\lineheight{1.25}\smash{\begin{tabular}[t]{l}(b)\end{tabular}}}}%
    \put(0,0){\includegraphics[width=\unitlength,page=3]{underactuated_design.pdf}}%
  \end{picture}%
\endgroup%